# A Study on Human-Swarm Interaction: A Framework for Assessing Situation Awareness and Task Performance

Wasura D. Wattearachchi, Erandi Lakshika, *Senior Member, IEEE*, Kathryn Kasmarik, *Senior Member, IEEE*, and Michael Barlow

***Abstract*—This paper introduces a framework for human swarm interaction studies that measures situation awareness in dynamic environments. A tablet-based interface was developed for a user study by implementing the concepts introduced in the framework, where operators guided a robotic swarm in a single-target search task, marking hazardous cells unknown to the swarm. Both subjective and objective situation awareness measures were used, with task performance evaluated based on how close the robots were to the target. The framework enabled a structured investigation of the role of situation awareness in human swarm interaction, leading to key findings such as improved task performance across attempts, showing the interface was learnable, centroid active robot position proved to be a useful task performance metric for assessing situation awareness, perception and projection played a key role in task performance, highlighting their importance in interface design and objective situation awareness influenced both subjective situation awareness and task performance, emphasizing the need for interfaces that emphasise objective situation awareness. These findings validate our framework as a structured approach for integrating situation awareness concepts into human swarm interaction studies, offering a systematic way to assess situation awareness and task performance. The framework can be applied to other swarming studies to evaluate interface learnability, identify meaningful task performance metrics, and refine interface designs to enhance situation awareness, ultimately improving human swarm interaction in dynamic environments.**

## I. Introduction

THE increasing complexity of tasks in dynamic environments has highlighted the importance of effective Human-Swarm Interaction (HSI) as a critical area of research. Robotic swarms, characterized by their decentralized and self-organized nature, operate independently based on local information [1]. While these systems demonstrate robust capabilities in autonomous operations, their limited sensory capacities and battery power [2] necessitate human intervention to enhance their performance. The concept of human-swarm teaming allows operators to guide robotic swarms without compromising their autonomy, effectively bridging the gap between human decision-making and robotic execution [3].

Despite advances in HSI research, significant gaps persist in the understanding and integration of Situation Awareness (SA) within these interactions. SA refers to an operator's ability to perceive, comprehend, and anticipate the state of the environment and the actions of the swarm within it [4]. Maintaining an appropriate level of SA is crucial for enabling operators to make informed decisions and effectively direct the swarm. Prior studies have often overlooked explicit SA measurement, assuming a direct relationship between SA and task performance (TP). This assumption can be misleading, as high levels of SA do not always guarantee successful task outcomes [5]. Furthermore, existing frameworks and principles for effective HSI often lack systematic approaches for incorporating SA, particularly in the context of dynamic environments where uncertainties abound.

The contributions of this paper are:

- A framework for incorporating the SA concept in HSI interfaces and user studies.
- An implementation of the concepts of the proposed framework to design an interface for HSI in dynamic environments, incorporating ten principles that consolidate existing concepts from human robot interaction (HRI) and swarm intelligence.
- A user study analysing the relationship between SA and TP to identify SA-sensitive metrics and assess human-swarm interactions effectively.

Using this framework, our study yielded several key findings: (1) the HSI interface was learnable, as participants improved TP across attempts while maintaining stable SA; (2) centroid active robot position is a reliable TP metric for assessing SA, reflecting how users guided the swarm without disrupting its autonomy; (3) perception and projection (SA Levels 1 and 3) played a key role in TP, emphasizing their importance in interface design; and (4) objective SA had a correlation with both subjective SA and TP, reinforcing the need for interfaces that support real-time awareness (objective SA). Furthermore, the study highlights the framework's adaptability for evaluating SA in various HSI scenarios. By identifying key SA-sensitive performance metrics and informing interface improvements, this approach can be extended to other swarm-based applications to enhance human-swarm collaboration in dynamic environments. In another paper [6], we explored the TP of the participants in terms of proximity to the target and the robot deactivations to understand the effectiveness of the interface and the principles that we derived, which will not be discussed in depth in this paper.

The remainder of this paper is organized as follows. Section II provides an overview of the background and related work. Section III introduces the proposed framework, while Section IV outlines one implementation of our framework as an HSI interface and user study. Section V presents the results of experiments using the designed interface and user study. Finally, the paper concludes with a discussion of the findings and suggestions for future work in Section VI.

## II. Background and Related Work

This section explores the foundational theories surrounding HSI and SA, the significance of SA measurement and the relationship between TP and SA, emphasizing their importance in effective HSI.

### *A. Human-Swarm Teaming*

Robotic swarms are decentralized and self-organized, with each robot operating independently based on local information, without the need for a central controller. The overall behaviour of the swarm emerges from the interactions and cooperation between individual robots [3]. Due to limited sensory and battery power [2], human intervention can become crucial. Through human-swarm teaming or HSI, operators can guide the swarm without compromising its autonomy [3].

HSI techniques can be categorized based on influence, communication mechanisms, and observation techniques. Influence-based interaction can be direct or indirect. In direct interaction, the human operator intentionally influences swarm behaviour, by selecting algorithms, controlling leaders, or manipulating swarm members through avatars [3, 7, 8]. Indirect interaction involves manipulating variables [9] such as swarm radius, inter-robot distance, environmental factors (e.g., virtual pheromones [10], beacons [11], breadcrumbs [12]) or coarse instructions [13]. Communication-based interaction can be remote or proximal. Remote interaction, such as teleoperation through graphical user interfaces (GUIs) [7] or virtual reality [21], is commonly used in high-risk environments [3]. Proximal interaction allows for more direct involvement, such as using augmented reality to manipulate virtual objects, gesture recognition, or speech commands [4], [14], [15].

Observation techniques are vital for decision-making, with past studies exploring swarm visualizations like individual robots [16], sub-teams [17], and area-based views [18]. Metrics such as alignment, grouping, and coverage help summarize swarm information for effective task management [19],[20].

In this paper, humans interact with the robotic swarm via a tablet-based interface, using remote and mixed methods. Indirect interaction involves marking grid cells to avoid, while direct interaction uses swipes to guide robots within a neighbourhood radius.

### *B. Road to Effective Interactions*

Goodrich and Olsen proposed seven principles for designing Multi-Robot Systems (MRSs) [21], but not HSI. These principles include tacit mode changes, natural signals, manipulating the environment instead of robots, controlling robot-world associations, interacting with interface information, reducing memory demands, and managing attention. Additionally, ten features for HSI have been suggested [8], such as fostering emergent intelligence, supporting local interactions, scalability, autonomous operation with human input, multiple users, enhancing SA, and user-friendly interfaces. While these features guide HSI, achieving efficient interaction is still challenging. Moreover, several taxonomies for MRS [22] focus on autonomy levels, human-robot ratios, decision support, task criticality, and robot morphology, primarily addressing "what" to design, not "how". Combining swarm-specific features with the Goodrich and Olsen model can lead to more effective interaction strategies. This paper introduces ten principles that integrate these concepts, existing HRI taxonomies, and swarm features to enhance HSI planning.

### *C. Situation Awareness*

Endsley [23] defines SA as *"the perception of elements in the environment within a volume of time and space, the comprehension of their meaning and the projection of their status in the near future"* by dividing SA into three levels: Level 1-Perception (L1); Level 2-Comprehension (L2); and Level 3-Projection (L3). To identify the SA requirements for a particular task, a technique called Goal-Directed Task Analysis (GDTA) [24] was introduced and has been adapted for various fields, including medicine [25], maritime operations [26] and firefighting [27].

Subjective and objective SA measurements are key methods for evaluating SA. The most popular technique for subjective SA is the Situation Awareness Rating Technique (SART) [28], which relies on participants' self-assessment to gauge their awareness levels, often using post-trial ratings. In SART, participants rate their SA across ten generic constructs (ten questions) across three main construct domains: Demands on Attentional Resources, $D$ (instability $I$, complexity, variability); Supply of Attentional Resources, $S$ (arousal, concentration, division of attention, spare capacity); and Understanding of the Situation, $U$ (information quantity, information quality, familiarity $F$) [28]. A seven-point rating scale (1 = Low, 7 = High) can be used [28] to gather ratings. The overall *SART Score* ($S_{SART}$) is calculated using Equation (1):

$$S_{SART} = U - (D - S) \quad (1)$$

In contrast, the Situation Awareness Global Assessment Technique (SAGAT) is the most commonly used objective method [5], [4], where a task is paused and blacked out, and participants queried about their SA as defined by Endsley's three levels [23]. Here onwards, the overall SA measured using SAGAT will be referred to as $S_{SAGAT}$.

Both techniques offer valuable insights: SART captures personal perceptions, while SAGAT assesses factual awareness. Both are used in our research to compare both.

### *D. Relationship between Situation Awareness and Task Performance*

The interplay between SA and TP is complex and multifaceted. Research has highlighted that a decline in SA can negatively impact TP. For instance, a lack of sensory information regarding the swarm's status has been shown to diminish SA, subsequently reducing TP [7].

Maintaining an appropriate level of SA is crucial not only for achieving high TP but also for informed decision-making and understanding the causes of different situations. For instance, even when TP is suboptimal, inadequate SA can hinder operators' ability to identify system failures and their causes. A study on pilots using SAGAT found that despite improved SA in a new fighter aircraft, mission performance was still suboptimal. This was due to pilots not fully adapting their tactics to the enhanced SA and the lack of performance metrics suited to combat scenarios [5]. These findings show that SA is vital for decision-making and system adaptability, beyond just TP. While this study focuses on SA's impact on TP, broader implications are left for future work.

### *E. Situation Awareness in Human-Swarm Interaction, Human-Robot Interaction and Dynamic Environments Research*

Several studies have explored SA in HSI, HRI and MRS, considering human interaction, swarming, explicit SA measurement, the relationship between SA and TP and dynamic environments. A study on HRI in urban search and rescue assessed SA by having users sketch locations post-experiment [29]. However, it was not conducted in a swarming setting and did not use standard SA measures. Several studies that involved both human interaction and swarming without considering SA aspects, including controlling swarm members using avatars [8], utilising physiological measurements in an adversarial task with HSI [30], observing swarm metrics related to the breadcrumbs model on a hand-held device [12], mixed granularity studies with HSI [31], a game benchmark to assess an HSI task [32], and using GUI, LEDs, and audio from robots for HSI [11]. On the other hand, studying the effect of latency and information loss on SA in teleoperated systems [17] explicitly measured SA using electroencephalography data and SAGAT, but did not explore its relationship with TP. Another study, measuring human swarm teaming performance with limited SA [7], considered SA in relation to TP through leader selection, but did not employ a standard method for SA measurement. Other studies that did not involve swarming considered dynamic environments. A study of humans-agent collaboration to guide first responders in a disaster scenario [33] examined an emergency response to a crashed radioactive satellite, and investigated the impact of latency and automation on teleoperation performance and trust in unmanned vehicle systems [34]. To the best of our knowledge, no prior studies have explored human interaction in swarming while explicitly measuring SA using a standard method, considering its relationship with TP, and conducting the study in such a dynamic environment.

Despite advancements in HSI and SA, significant gaps remain. Many studies focus on TP without explicitly measuring SA [7, 35], assuming a direct correlation that does not always hold. Systematic approaches for conducting SA studies in HSI are lacking. Simply presenting information does not ensure accurate perception, comprehension, or projection of the situation unless SA is explicitly measured. Although established methods exist for HRI and SA in HRI [21, 36], there is an absence of specific methodologies tailored to HSI with SA. Furthermore, research on HSI in dynamic environments is limited, with many studies neglecting the uncertainties affecting SA and TP, especially involving low-cost, sensory-limited robotic swarms. This study addresses these gaps by integrating SA measurement into HSI design, optimizing TP in dynamic environments, and improving human-swarm teaming under real-world conditions. Our framework for this integration is presented in the next section.

## III. Proposed Framework For Assessing Situation Awareness and Task Performance

Our framework proposed here contains four main steps as shown in the Unified Modelling Language activity diagram in Fig. 1. The steps are: Do SA requirement analysis; do SA measurement design; do interaction design; and do user study. Each of these steps will be discussed in detail in this section.

### *A. Step 1: Do Situation Awareness Requirement Analysis*

The first step is to define a swarming task, such as navigation, target search, adversarial patrolling or any other task that utilizes a robotic swarm to achieve a specific goal. After identifying the task, a GDTA [24] is conducted to determine overall goals, sub-goals, key decisions, which leads to the identification of the necessary SA requirements categorised into Endsley's three levels of SA. This categorization is particularly valuable for doing SA measurement in Step 2. Next, the SA requirements should be grouped into information dimensions which serve as abstract representations of SA requirements, categorized based on the specific type of knowledge needed for the task. This helps organize SA requirements and supports interface design in Step 3, ensuring that the user interface is developed in a way that facilitates SA and TP. The outputs of this step are Goals, Sub-goals, Decisions, SA Requirements (categorized by levels), and Information Dimensions, where the latter two will be inputs to Steps 2 and 3 in order.

### *B. Step 2: Do Situation Awareness Measurement Design*

SA requirements from Step 1 are used to design SAGAT queries [4, 5] for the swarming task, to capture the objective SA considering the three SA levels.

### *C. Step 3: Do Interaction Design*

Interaction design can be performed in parallel with measurement design (Step 2). This process requires two key inputs: the information dimensions identified in Step 1 and the ten principles of efficient HSI derived from previous literature [6]. Ten principles have been introduced in this research that combine the concepts from Goodrich and Olsen [21] and existing taxonomies for HRI [22], with swarm features to be more robust when planning HSI [8] as listed below:

1. Utilise local information and create a global picture out of it
2. Facilitate switching between swarm and human control
3. Use versatile interaction methods

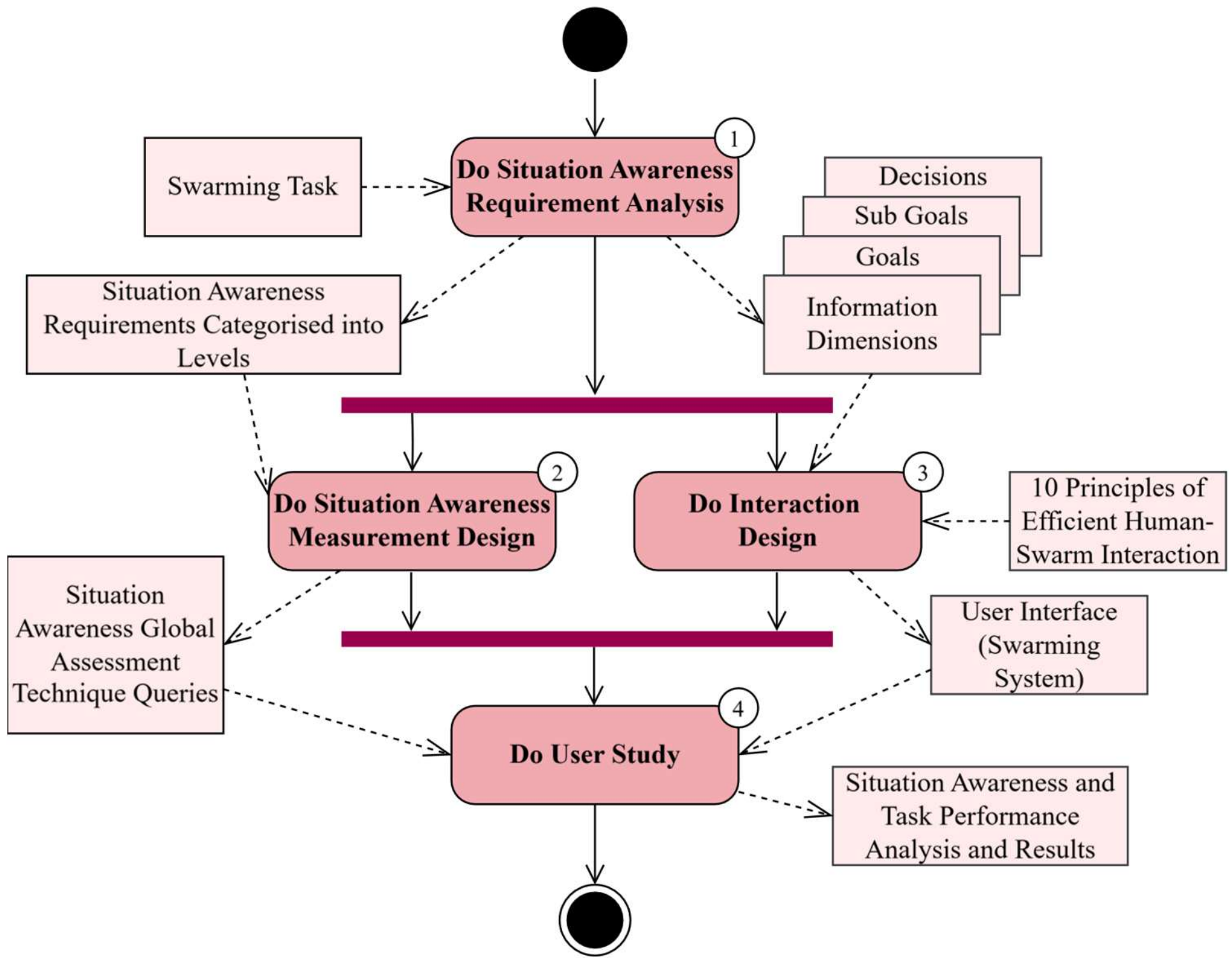

**Fig. 1.** Proposed framework for integrating SA in HSI.

4. Tacitly switch between interaction modes, methods and tools
5. Use intrinsic modalities over conventional methods and interfaces
6. Minimise the interaction gap between the swarm member(s) and the human during direct interaction
7. Minimise the interaction gap between the environment and the human during indirect interaction
8. Enable engagement with the information within the interface
9. Reduce what should be remembered
10. Assistance in managing the attention

Applying these principles to the information dimensions helps define the necessary interactions for designing an effective user interface for the swarming system.

### *D. Step 4: Do User Study*

The final step involves a user study to measure both SA and TP, which helps identify relationships between them and key SA elements influencing TP. The primary input for this step is the user interface, with SAGAT queries assessing objective SA during the task. Additionally, TP can be measured alongside SA to provide a comprehensive evaluation. Meanwhile, other standard techniques of measuring SA (for example, SART for subjective SA) also can be incorporated. The final output includes analysis and results that will inform potential improvements to the interface, enhancing SA and TP.

## IV. Applying the Framework to Build a User Interface and Conducting a User Study

This section provides an example application of the framework proposed in Section III.

### *A. Do Situation Awareness Requirement Analysis*

**Swarming Task:** Particle swarm optimisation (PSO) [37] is one important swarming approach that has been applied in robotics as it is useful for tasks such as locating a chemical, biological, radioactive or nuclear (CBRN) hazard target [8],[38],[39]. Thus, in this paper, a single target search was designed where 20 robots executed PSO in an open environment that contains both static and dynamic obstacles (which are designated as regions to avoid). The primary objective is to have multiple robots successfully locate the target and avoid entering restricted regions (which leads to robots becoming damaged/deactivated). The robots employ the PSO LBEST model [40] with a limited communication range and incorporate a basic obstacle avoidance algorithm. They can detect a virtual signal emitted by the target source modelled by the fitness function in Equation 2 (from [41]), calculating the intensity $\beta$ of a constant source of power $P$ at a given Euclidean distance $d$ to the robot. The objective of PSO is to locate the maximum of this fitness function, thus directing robots toward the power source target.

$$\beta = P/d^2 \tag{2}$$

We implemented the robots in simulation using CoppeliaSim Human operators assist the simulated robotic swarm using a tablet running an Android application GUI which will be discussed more on Section IV-C. They receive messages about hazardous regions and can mark areas to avoid on the map. The type and pattern of these alerts define three task variations: distributed (Dis), moving (Mov), and spreading (Spr) hazards. In the Dis scenario, random grid cells become hazardous at intervals, simulating threats such as rough terrain, strong winds, weak signals, or adversarial camps. The Mov scenario represents a shifting threat, where hazardous cells become safe as the hazard moves to adjacent locations, modelling adversarial patrolling. In the Spr scenario, the threat originates in one cell and expands to neighbouring cells, simulating events like fire or contamination. By comparing these scenarios, we aim to analyse TP across different hazard types and assess if the findings are generalizable. While robots can locate targets independently, human operators are responsible for minimizing deactivations from robots entering hazardous areas.

**Applying GDTA to the swarming task and forming information dimensions:** As detailed in Section III, GDTA was performed for the swarming task, identifying SA requirements and information dimensions as shown in Table I. Each SA requirement is labelled as DimX.Y in the table, where X represents the information dimension number and Y is the list number within that dimension. Six dimensions were derived: Dim 1: Location information of active robots and the target, Dim 2: Motion and spatial state information of the robots, Dim 3: Temporal progress information, Dim 4: Dynamic obstacles status information, Dim 5: Robot loss information and Dim 6: Robot trapped information.

*B. Do Situation Awareness Measurement Design*

SAGAT queries were developed to assess the SA requirements in Table I and will be presented to the user by pausing the task periodically. Additionally, an online standard SART questionnaire [28] used to assess subjective SA at the end of each task.

*C. Do Interaction Design*

The proposed ten principles were applied to the target-finding swarming task that was discussed in Section IV-A and a tablet-based interface has been developed as shown in Fig. **2** using Android Studio. Users interact with the system in two ways: tapping cells to mark hazard zones (in red) or long-tapping to remove them and swiping to apply force to nearby robots within a predefined radius, moving them in the direction of the swipe. A timer in the top right shows the remaining task time.

*D. Do User Study*

The final step in the methodology is a user study that evaluates participants' SA using designated measurement techniques and TP primarily through the centroid active robot position (CA). Additional metrics, such as the nearest active robot position (NA) and the nearest quartiles active robot positions (Q1 - NAQ1, Q2 - NAQ2), were also used when necessary. The user study was conducted with 31 participants (15 males, 15 females, 1 preferred not to disclose). The study followed the procedure outlined below.

**Consent:** First, consent was obtained from participants, confirming they were over 18 years old and comfortable with reading brief abstract text messages related to fire, falling objects, or natural disasters (e.g., rocks falling or strong winds).

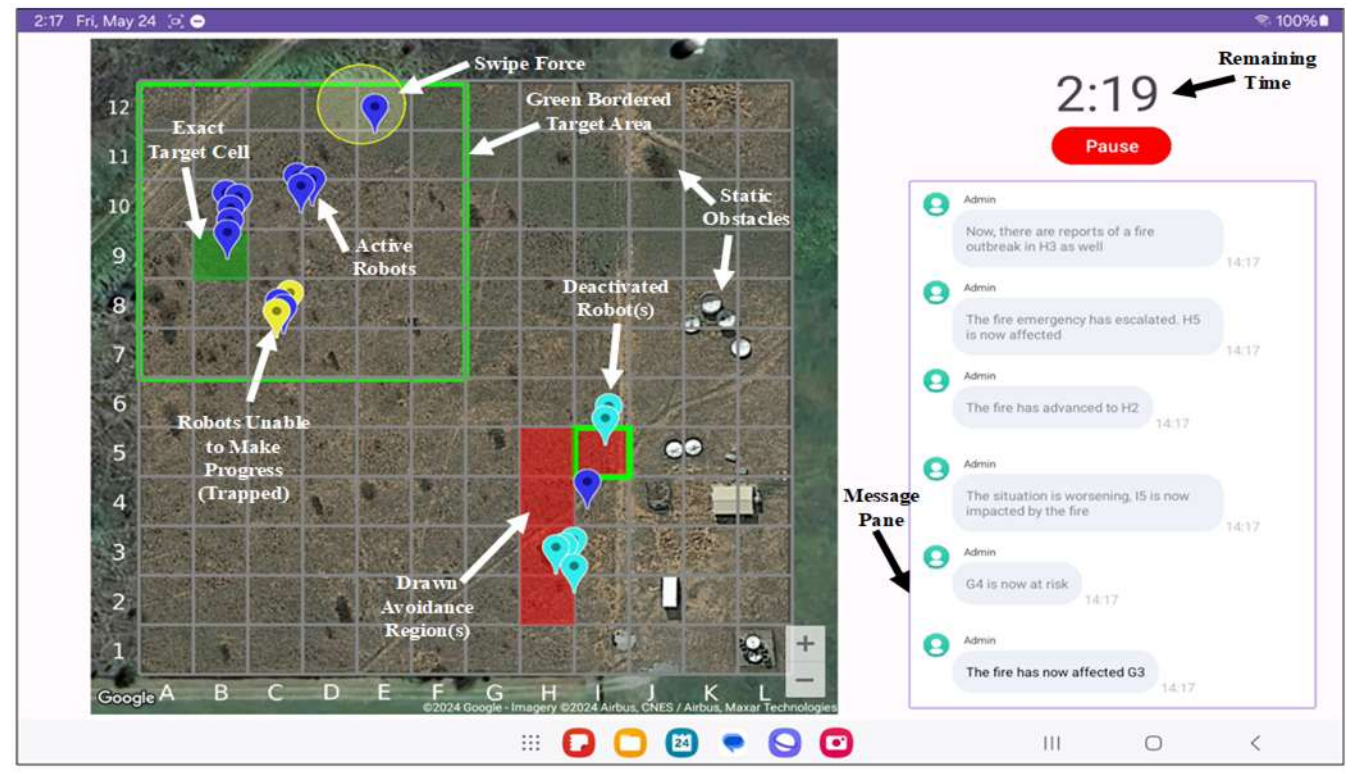

**Fig. 2.** The designed tablet interface.

**Pre-task questionnaire:** Participants completed a questionnaire via Qualtrics to provide demographic information such as age, gender, highest educational level, familiarity with touch-based devices, and familiarity with robotics swarms. Further details are provided in the supplementary material Section I.

**Instructional video:** Participants watched a 10-minute video explaining the tasks and how to use the tablet interface, followed by a question-and-answer session with the facilitator.

**Task execution:** Participants used a tablet with the GUI described in Section IV-C and were assigned tasks based on one of the three hazard types, with five minutes to complete each task. During the task, two pauses occurred, hiding the screen, during which participants answered 28 SAGAT questions (14 per pause) to assess objective SA. They were instructed to respond quickly to focus on SA rather than memory recall. The assessment included Multiple Choice Questions (MCQs) and Cell Marking Questions (CMQs). For MCQs, participants chose one of five options, while CMQs required them to mark cells on a grid, similar to marking avoidance regions during the task. The survey proceeded sequentially without a back button. As explained in the instructional video, if participants were unsure of an answer, they were explicitly instructed to select "I don't know" rather than guess. For grid marking questions, a "Not applicable" option was available when no correct answers were possible, ensuring accurate scoring.

**Post-task questionnaire (SART):** At the end of each task, participants completed an online questionnaire via Qualtrics to assess their subjective SA comprised 10 questions, corresponding to the 10D SART [28].

**Task repetition and attempts:** After completing a task for a specific hazard type, participants proceeded to another two tasks for the other hazard types, following the same procedure. This sequence was termed 'Attempt 1 (A1)'. The three tasks were then repeated in a different order as 'Attempt 2 (A2)'. Each task within an attempt is referred to as a 'participant-task', with each participant completing six participant-tasks in total. The study was completed within two hours.

## VI. Experiments, Results and Analysis

This section presents the analysis and results of the study, focusing on four experiments. First, it explores the learnability of our HSI system using SA by examining whether there are any significant differences between A1 and A2. Second, it discusses the way of identifying a TP metric to assess SA directly. All the results are reported based on 95% confidence level and $p < 0.05$ was considered as significant. Then, the third and fourth experiments were conducted to identify correlations between objective SA, subjective SA, their elements and TP.

TABLE I
GOAL-DIRECTED TASK ANALYSIS FOR A TARGET SEARCH IN A DYNAMIC ENVIRONMENT USING A ROBOTIC SWARM

| Goals | Sub-goals | Decisions | SA Requirement |
|---|---|---|---|
| Find the target within the given time | Move active robots as much as closer to the target | How much closer are the active robots to the target?<br>How much time is left?<br>How to let swarm autonomy carry out the active robots near the target (converge)? | ***L1:***<br>Dim1.1 Positions of the active robots<br>Dim1.2 Position of the target<br>Dim3.1 Remaining time to finish the task<br>***L2:***<br>Dim1.3 Distance between the target and the active robots<br>Dim1.4 Number of active robots near the target<br>Dim3.2 Time spent on the task<br>Dim1.5 Region where the most activated robots are located<br>Dim2.1 Speed of the robots<br>Dim2.2 Robots' direction of movement<br>Dim2.3 Spatial distribution of the robots across the search space<br>Dim2.4 Spatial distribution of the robots with respect to the target<br>***L3:***<br>Dim1.6 Number of active robots that can come near the target in the next few seconds<br>Dim2.5 Spatial distribution of the robots across the search space after a few seconds<br>Dim2.6 Spatial distribution of the robots with respect to the target after a few seconds<br>Dim3.3 Time needed to complete the task |
| | Move many active robots closer to the target | How many active robots are closer to the target?<br>How to guide the automated swarm toward the target?<br>Will the remaining time be enough to send many active robots near the target? | |
| Minimize the damage to the robots | Avoid robots entering hazardous regions (dynamic obstacles) | How many hazardous regions (dynamic obstacles) are currently active?<br>How many robots are deactivated after entering hazardous regions?<br>Can the operator avoid robots entering a particularly hazardous region? | ***L1:***<br>Dim4.1 Positions of the active hazardous regions<br>Dim4.2 Positions of the drawn regions to avoid<br>Dim5.1 Positions of the deactivated robots<br>***L2:***<br>Dim5.2 Region where the most deactivated robots are located<br>***L3:***<br>Dim4.3 Number of hazardous regions that might appear within the next few seconds<br>Dim4.4 Number of drawn hazardous regions that can be removed within the next few seconds<br>Dim5.3 Number of robots that can be deactivated next few seconds |
| | Avoid robots clashing or getting trapped | Are there any robots trapped in between static obstacles?<br><br>How to move/assist trapped robots? | ***L1:***<br>Dim6.1 Positions of robots that are trapped between obstacles and unable to move<br>***L2:***<br>Dim6.2 Region where the most trapped robots are located<br>***L3:***<br>Dim6.3 Number of robots that can get trapped in the next few seconds |

*A. Experiment 1: Evaluating the learnability of our Human-Swarm Interaction System using Situation Awareness*

**Aim:** To determine whether the HSI interface designed by incorporating SA concepts is learnable.

**Hypothesis:** The interface is learnable as indicated by significantly improved TP (denoted by decreased CA) and *U* between A1 and A2.

**Method:** We analysed data collected from our user study outlined in Section IV, for each hazard separately. As mentioned in Section IV-D, TP was assessed using CA (mainly). For each participant-task, scores were calculated for each construct, along with $S_{SART}$ obtained by averaging all responses. Similarly, for objective SA, questions were grouped by SA levels and information dimensions. For each participant-task, the average score was computed for each SA level, each information dimension, and $S_{SAGAT}$ by averaging all responses (each scored out of 100). Then Wilcoxon paired tests were conducted to determine statistical significance in improvements for TP, subjective SA and objective SA scores.

**Results:** As shown in Table II, the CA significantly decreased in Dis (improved TP) and increased in Mov (declined TP). No significant CA difference was observed for Spr ($p = 0.14$). Additional results on the TP metrics can be found in the supplementary material Section II-B. When considering Subjective SA, *U* increased for Dis and Spr hazards (driven by *F*), while the *D* increased significantly for Mov mainly due to *I*. $S_{SART}$ did not show significant difference between attempts. Objective SA elements showed mixed results (can be found in the supplementary material Section II-B) and $S_{SAGAT}$ did not show any significance across attempts.

**Conclusion:** Overall, TP improved from A1 to A2 for Dis, showed no significant difference for Spr, and declined for Mov. Participants' *U* in subjective SA increased for Dis and Spr, but not for Mov. Objective SA results were mixed and did not reveal clear trends. There were no significant differences between attempts for $S_{SART}$ and $S_{SAGAT}$. These findings suggest that increased *U* in Dis and Spr was linked to improved or consistent TP, supporting the hypothesis that a learnable HSI leads to improved TP.

*B. Experiment 2: Identifying a task performance metric for assessing Situation Awareness*

**Aim:** To identify a TP metric that reliably reflects SA in HSI.

TABLE II
SIGNIFICANCES OF TASK PERFORMANCE (CA) AND SUBJECTIVE SA ($U$, $F$, $D$, $I$) RESULTS BY HAZARD TYPE

| Hazard | Metric | Attempt 1 Median [IQR] | Attempt 2 Median [IQR] | Significance |
|---|---|---|---|---|
| **Dis** | CA (m) | 7.8 [5.2 - 13.1] | 2.5 [1.1 - 9.5] | Yes |
| | $U$ | 14.0 [14.0 - 16.0] | 16.0 [14.5 - 18.0] | Yes |
| | $F$ | 4.0 [3.5 - 5.0] | 5.0 [4.50 - 6.0] | Yes |
| **Mov** | CA (m) | 2.4 [0.8 - 4.0] | 9.5 [4.5 - 20.6] | Yes |
| | $D$ | 11.0 [8.5 - 13.0] | 12.0 [10.0 - 15.0] | Yes |
| | $I$ | 4.0 [3.0 - 5.0] | 4.0 [3.5 - 6.0] | Yes |
| **Spr** | CA (m) | 18.3 [8.2 - 27.8] | 11.9 [9.2 - 19.3] | No |
| | $U$ | 15.0 [14.0 - 17.0] | 17.0 [15.0 - 18.0] | Yes |
| | $F$ | 5.0 [3.0 - 5.0] | 6.0 [5.0 - 6.0] | Yes |

**Hypothesis**: Since HSI should allow users to assist the swarm without disrupting its autonomy, a suitable TP metric should align with both effective swarm behaviour and the operator's SA. The hypothesis is, a TP metric that shows a meaningful relationship with SA scores, reliably assesses SA.

**Method:** The same TP metrics as in Experiment 1 were used, and correlations between CA and SART and SAGAT scores were analyzed, focusing on A2 due to more significant correlations and improved $U$ (as highlighted in Section VI-A) in this attempt. When assessing correlations, two types of scores were considered: the mean score across all tasks performed by a user in A2 (M_A2_all) and the score obtained by a user for a specific hazard type in A2 (S_Hazard_A2). Spearman's correlation assessed relationships, with significance at $p < 0.05$ and strength classified as strong ($\geq 0.70$), moderate (0.50 - 0.70), and weak ($< 0.50$). The closer the robots are to the target (i.e., the lower the TP score), the better the TP. Therefore, if higher TP and SA are expected, this should be reflected as negative correlations between TP metrics and SA scores.

**Results**: There was a moderate correlation (-0.61) when considering M_A2_all with $S_{SAGAT}$ and CA as shown in Fig. 3 (The orange lines in the figures are second-degree polynomial trends with a 95% confidence interval denoted by the gray area). In the analysis of different hazard types, only the Spr exhibited moderate correlations between the SAGAT and TP metrics, all of which were observed in A2. Specifically, in S_Spr_A2, the moderate correlation of -0.58 was observed between $S_{SAGAT}$ and the CA.

When examining the relationship between different SA levels and TP, moderate negative correlations were found again. When considering M_A2_all, L1 and the CA had a moderate negative correlation of -0.64, followed by L3 (-0.60) and L2 (-0.32, weak), as shown in Fig. 4. In the analysis of different hazard types, the Mov exhibited moderate correlation between the L1 and CA (-0.54) in S_Mov_A2 condition. For Spr, there was a moderate correlation between L3 in S_Spr_A2 with CA (-0.63) as shown in Fig. 5. For information dimensions, Spearman's correlation analysis revealed moderate negative correlations without a clear pattern in A2.

**Conclusion:** The CA meets the criterion that a TP metric capable of assessing SA should exhibit a meaningful relationship with SA, as it correlated with SA scores while also reflecting effective swarm guidance. Therefore, this metric can be considered a reliable indicator of SA in HSI, supporting prior studies that suggest performance metrics can be used to infer SA.

*C. Experiment 3: Identifying correlations between objective Situation Awareness and task performance*

**Aim**: To examine the relationship between objective SA along with its elements and TP in A2. Understanding which SA elements correlate with improved TP can help refine interface design to enhance SA and, consequently, TP.

**Hypothesis**: Objective SA, and its SA levels and information dimensions show correlations with CA.

**Method**: The same method used in Experiment 2 was used.

**Results**: A negative correlation was observed when considering M_A2_all with $S_{SAGAT}$ and CA in A2 (-0.61), confirming that as SA improved, TP increased. This trend was particularly evident for L1 (-0.57) and Dim 1 (-0.55), reinforcing that Perception of robot and target location information is critical for effective task execution. These findings are visualized in Fig. 6. For Spr specifically, correlations were found when considering S_Spr_A2 between $S_{SAGAT}$ and CA (-0.64) as well as L3 and CA (-0.63), indicating that higher Projection led to better TP in this scenario (Fig. 7).

**Conclusion**: The results support the hypothesis that the higher SA scores—especially in L1 and L3—correlate with improved TP, reinforcing the importance of these SA elements in HSI. Additionally, certain information dimensions such as Dim 1 emerged as significant contributors to performance. These insights suggest that enhancing interface design to support perception and projection—particularly through improving location information of the robots and the target —can lead to increased TP.

*D. Experiment 4: Identifying correlations between Objective Situation Awareness, Subjective Situation Awareness, and Task Performance*

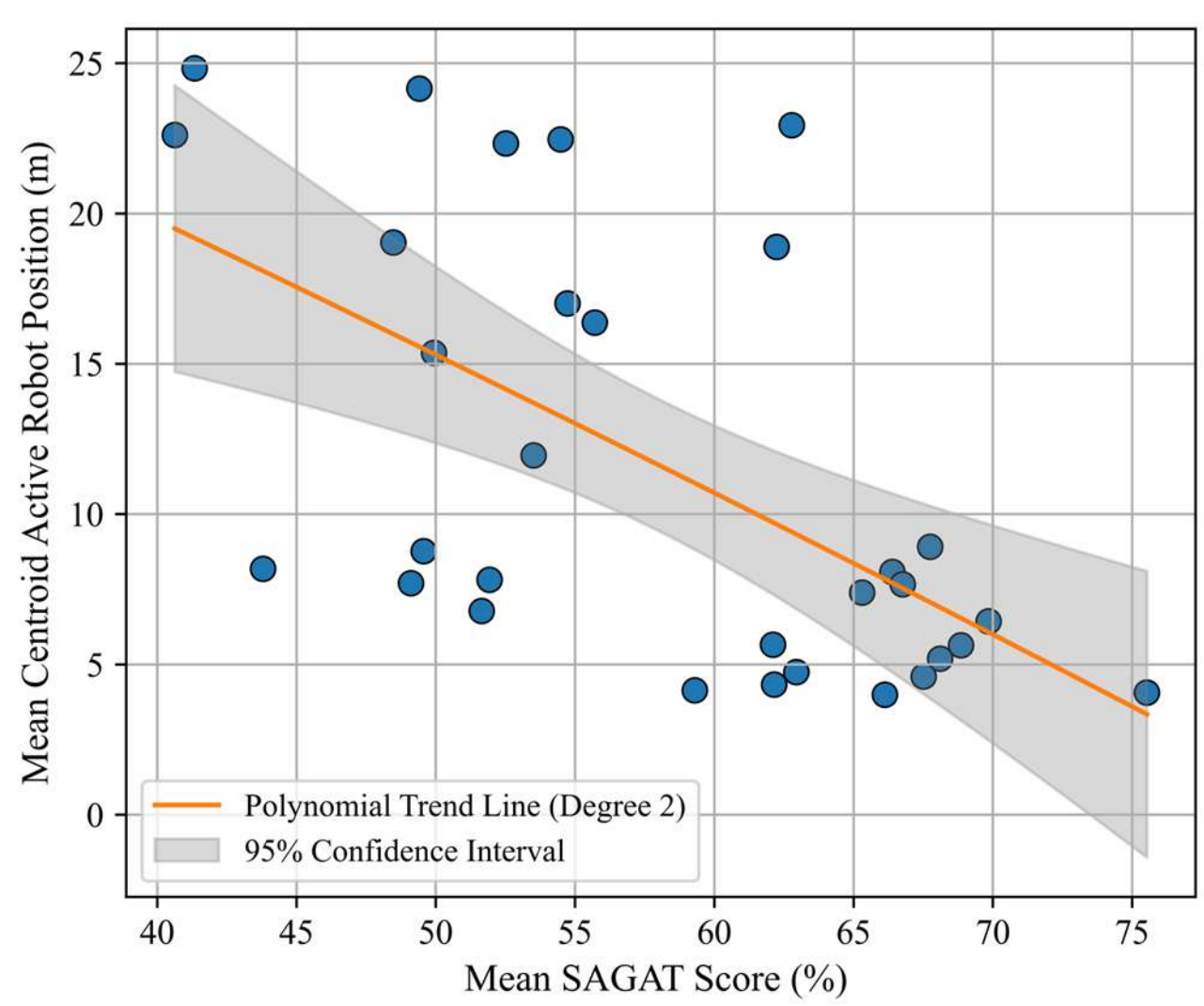

**Fig. 3.** The scatter plot showing the correlation between $S_{SAGAT}$ and CA for A2 across all tasks.

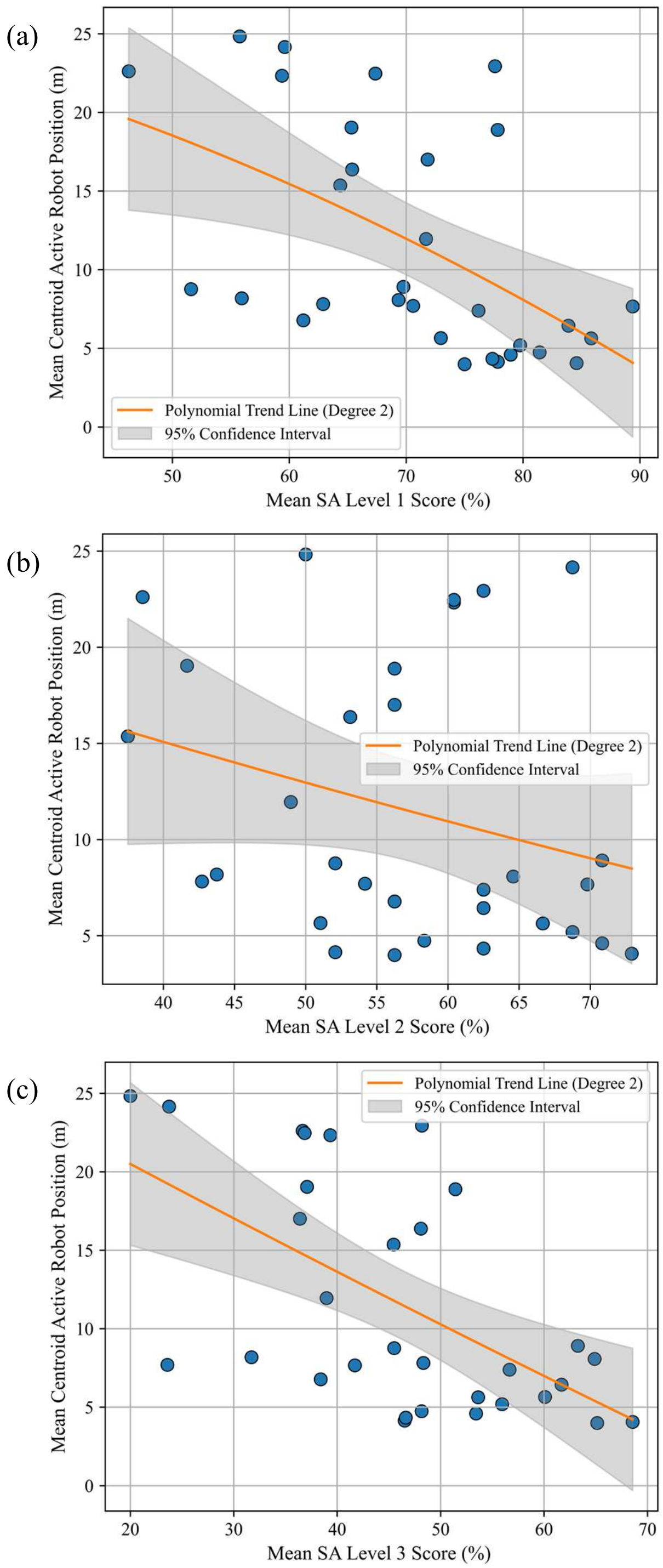

**Fig. 4.** The scatter plots showing the correlation between (a) L1; (b) L2; (c) L3 and CA for A2 across all tasks.

**Aim**: To investigate the relationship between objective SA, subjective SA, and TP in A2. Understanding whether both subjective and objective SA contribute to TP or not can provide insights into how SA should be supported in interface design.

**Hypothesis:** There are correlations between $S_{SAGAT}$, $S_{SART}$ and TP, as indicated by CA.

**Method**: The same method was used as in Experiments 2 and 3.

**Results**: Significant correlations were observed between $S_{SAGAT}$ and $S_{SART}$ (-0.51), and $S_{SAGAT}$ and CA (-0.61) but not with $S_{SART}$ and CA, when considering M_all_2. These results indicate that SA measured via real-time queries correlates well with both self-reported SA and TP, highlighting the value of objective SA assessment. The correlation patterns are visualized in Fig. 8, reinforcing the interplay between different SA measures and performance outcomes.

**Conclusion**: The results support the hypothesis that there are correlations between objective SA, subjective SA and TP. Participants who demonstrated higher SA based on measured accuracy (objective SA) also reported higher perceived awareness (subjective SA) and showed better performance. This suggests that interface designs that enhance objective SA can also lead to improved subjective SA and TP in HSI. Therefore, prioritizing elements that support objective SA is particularly important.

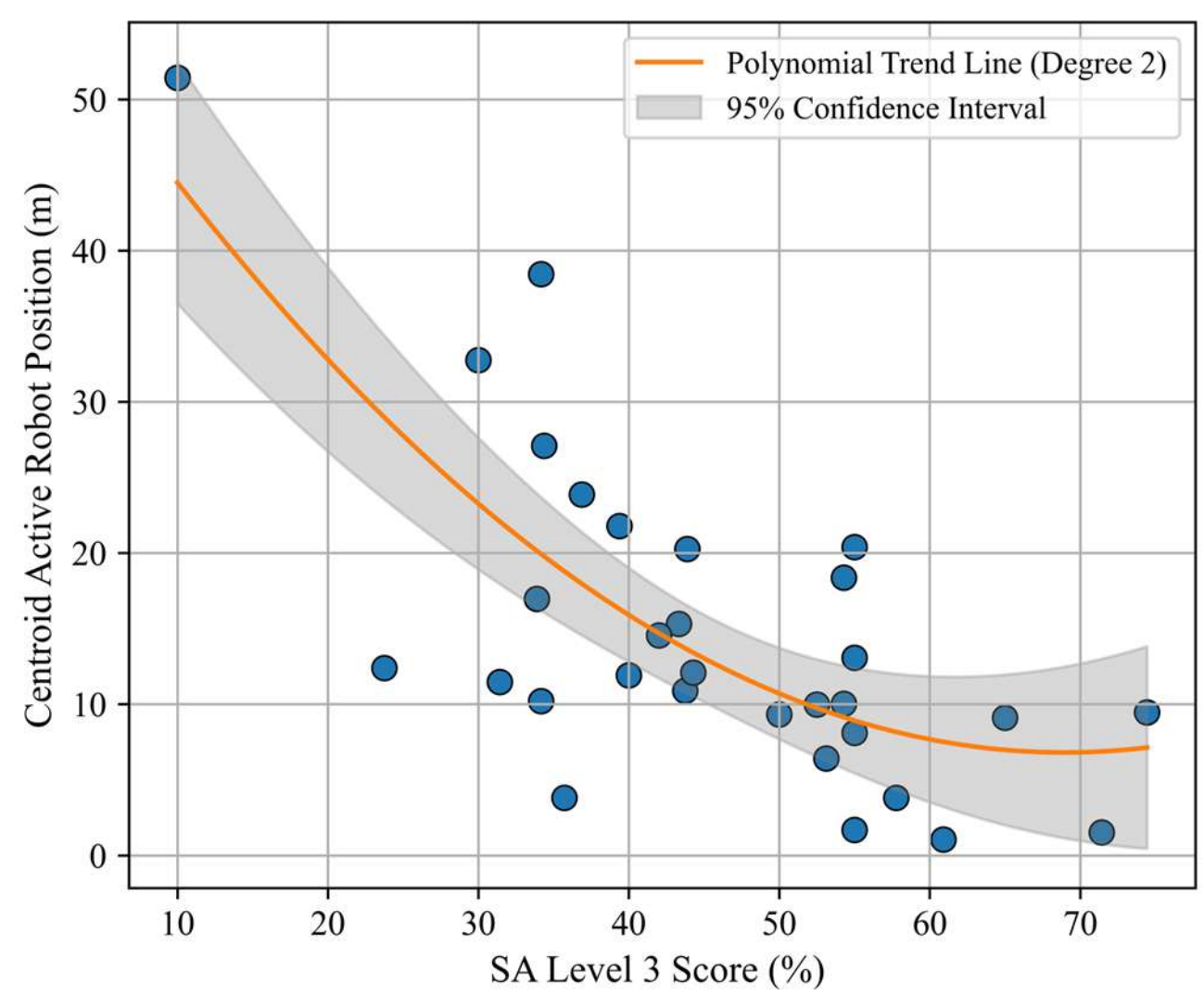

**Fig. 5.** The scatter plot showing the correlations between L3 and CA, for A2 in Spr.

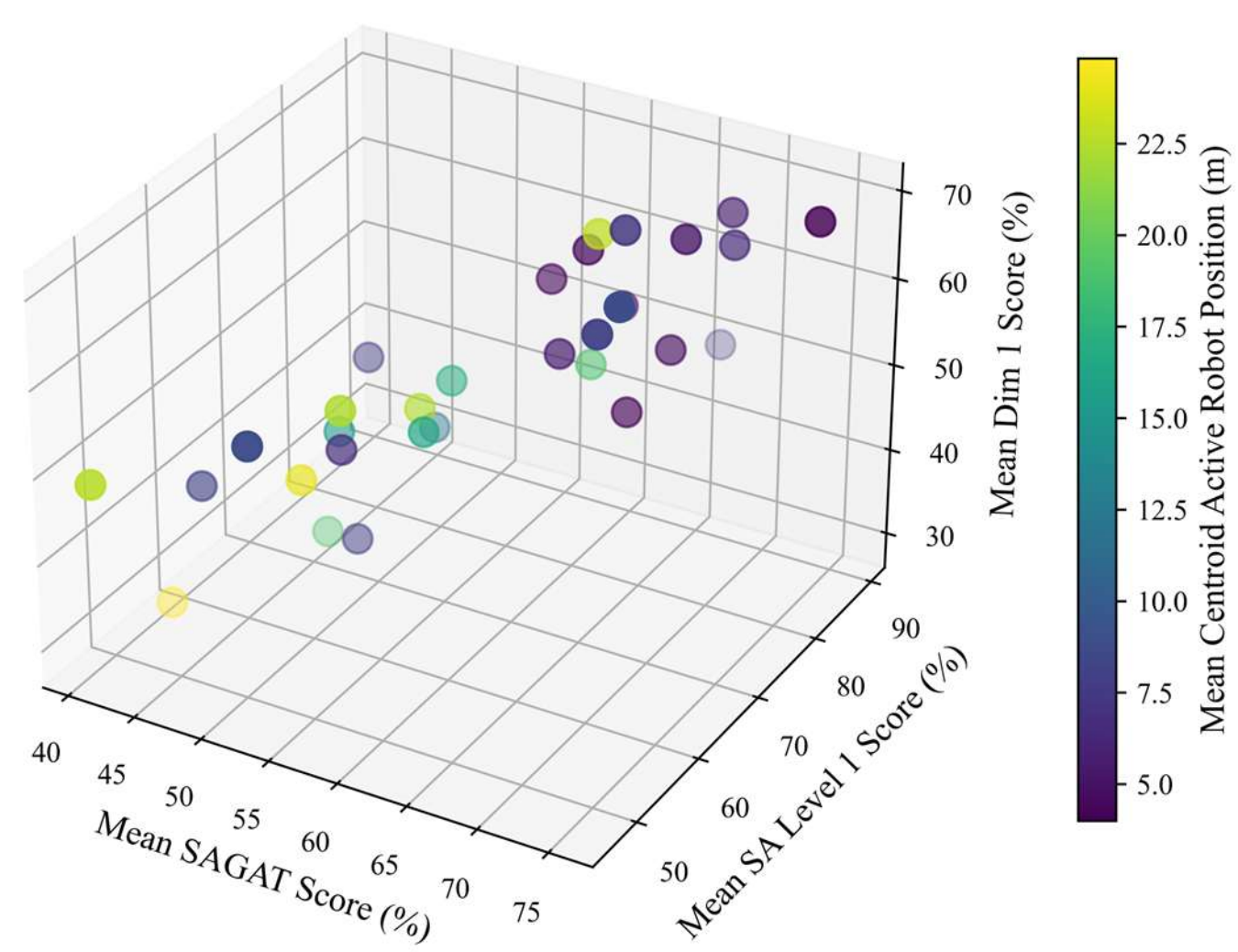

**Fig. 6.** Relationship between Mean SAGAT Score, L1, Dim 1, and CA in A2. The plot reveals that higher scores in these SA elements are linked to better TP (colour bar), indicated by a lower CA.

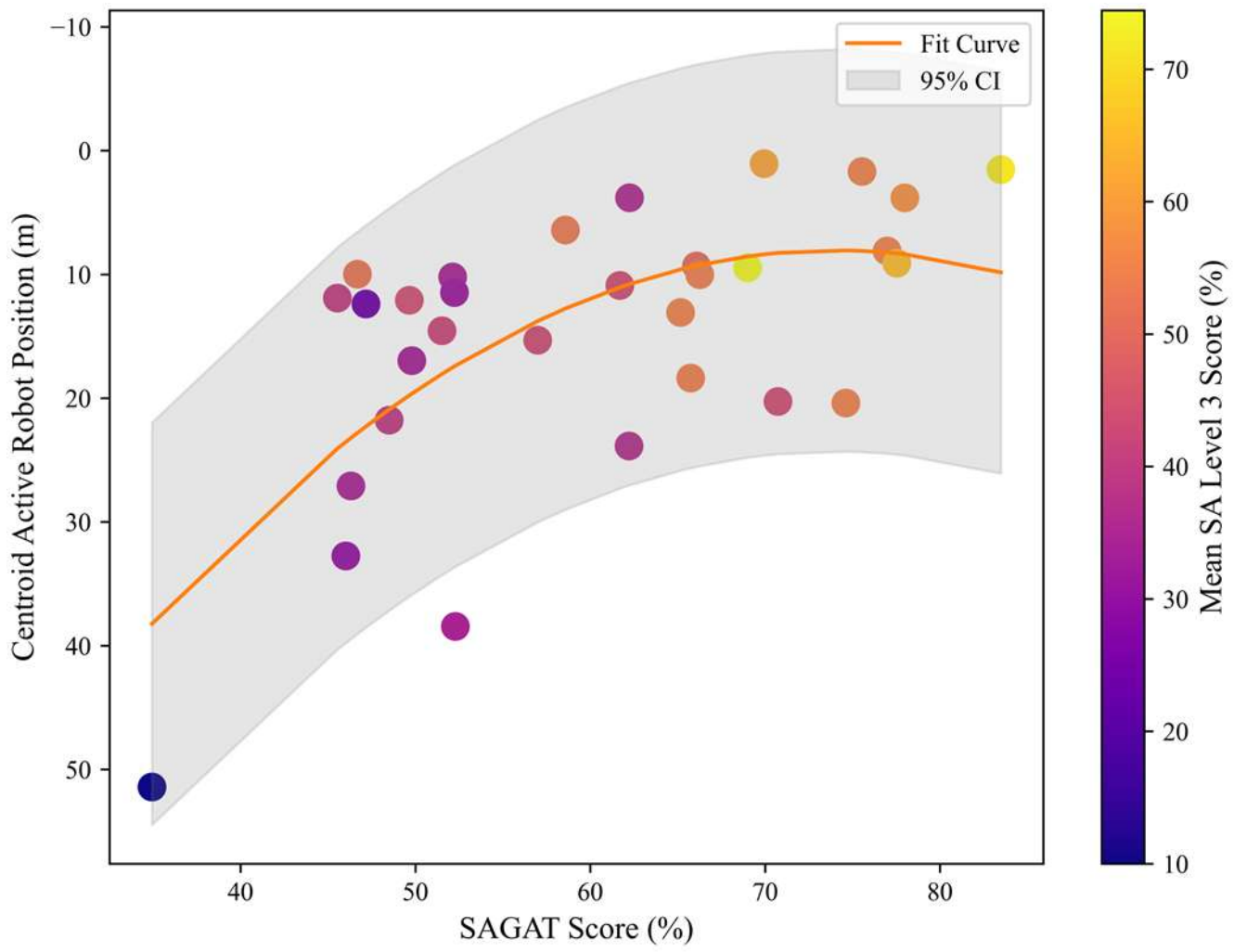

**Fig. 7.** Relationship between Mean SAGAT Score, L3, and CA in Spr during A2. This scatter plot highlights the relationship between participants' L3 (colour bar), $S_{SAGAT}$ (x-axis), and TP (y-axis), measured by the CA in Spr. The nonlinear fit curve suggests a complex interaction between these factors, where higher SA scores, particularly in L3, are associated with improved TP, as indicated by a lower CA.

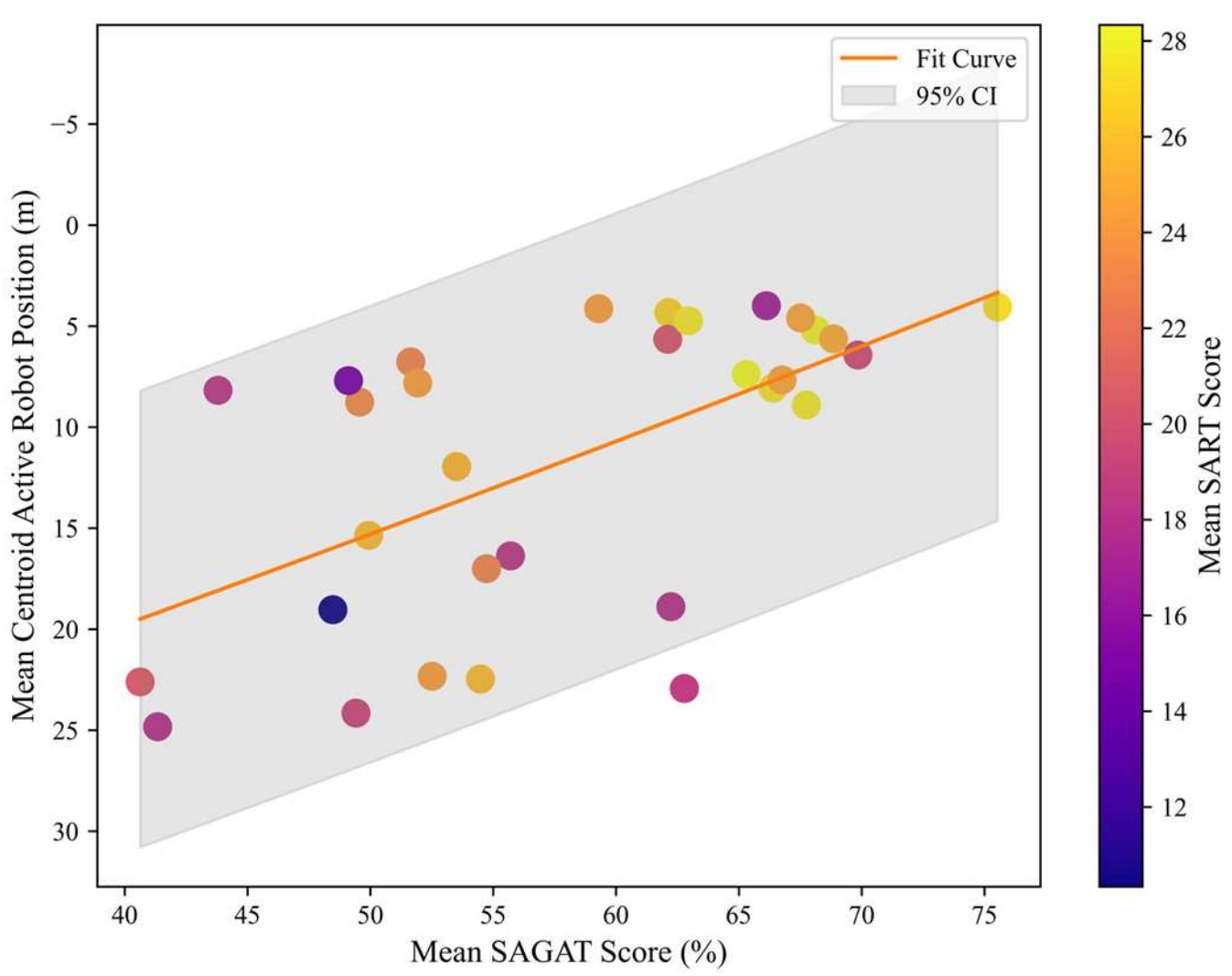

**Fig. 8.** Relationship between Mean SAGAT Score, Mean SART Score, and CA for all tasks in A2. This scatter plot illustrates the correlation between $S_{SAGAT}$ (x-axis), and $S_{SART}$ (colour bar), with TP (y-axis). The positive trend in the fit curve suggests that as both objective and subjective SA improve, the CA moves closer to the target, indicating enhanced TP.

## VII. Conclusions and Future Work

A comprehensive framework for conducting HSI studies incorporating SA was introduced, along with ten principles for effective HSI. A user study was conducted where a robotic swarm performed a single-target search assisted by a human participant using a tablet-based user interface to mark hazardous areas. The SA of human participants was evaluated both subjectively (using SART) and objectively (using SAGAT), alongside TP, which was assessed based on how close the robots were to the target at the end of the task. The tasks involved guiding robots under dynamic hazards categorized as Dis, Mov, and Spr.

This study investigated the learnability of a HSI interface through the lens of SA and explored the relationship between SA and TP. Four key analyses were conducted to assess interface learnability, identify a reliable TP metric for assessing SA, and determine the correlations between SA (both subjective and objective) and TP.

The findings demonstrated that, using our framework, the designed HSI was learnable. Participants' $U$ ($F$) improved between attempts while maintaining SA across subjective and objective measures. TP also increased in A2, supporting the hypothesis that a learnable interface leads to improved TP.

Additionally, our framework facilitated the identification of a TP metric that effectively reflects SA in HSI. Among various potential metrics, CA emerged as a reliable indicator, showing meaningful correlations with SA. These findings reinforce prior research suggesting that TP metrics can serve as indirect measures of SA.

Further analyses revealed that objective SA plays a crucial role in TP. Specifically, L1 and L3 correlated with improved TP, emphasizing the importance of enhancing these SA elements through interface design. A key information dimension, particularly location information of robots and the target, was also identified as significant contributors to TP, highlighting how improved interface support for these elements can enhance SA and overall task execution using our framework.

Finally, this study confirms that objective SA contributes to both subjective SA and TP. This highlights the importance of enhancing interface elements that support objective SA, as doing so can also positively influence users' perceived awareness and TP in HSI.

By applying our framework, this study demonstrates a structured approach to assessing SA and TP in HSI. The framework can be applied to other swarming studies to evaluate interface learnability, identify meaningful TP metrics, and refine interface designs that support SA, ultimately improving HSI in dynamic environments.

In future research, the role of task repetition and $F$ can be further explored to understand the extent to which these factors influence SA and performance. Additionally, refining the interface design by targeting specific SA elements like Dim 1 could help further improve TP. Expanding the study to other task scenarios and hazard types will also provide deeper insights into how these relationships evolve under different conditions.